\documentclass[runningheads]{llncs}
\usepackage{colortbl}
\usepackage{diagbox}
\usepackage{threeparttable}
\usepackage[T1]{fontenc}
\usepackage{graphicx}
\usepackage{tabularray}
\UseTblrLibrary{diagbox}
\usepackage[normalem]{ulem}
\usepackage{xcolor}
\usepackage[table,xcdraw]{xcolor}
\usepackage{graphicx,wrapfig,lipsum}
\usepackage{float}
\usepackage{amsmath}
\usepackage{afterpage}
\usepackage{multicol}
\usepackage{hyperref}

\usepackage{subcaption}

\usepackage{enumitem}
\begin{document}

\title{Weaknesses of Facial Emotion Recognition Systems}

\author{
Aleksandra Jamróz$^*$\inst{1}
\and
Patrycja Wysocka$^*$\inst{1,2}
\and
Piotr Garbat \inst{1}
}

\institute{
Warsaw University of Technology, Poland
\and
SWPS University, Poland \\
\email{oyamroz@gmail.com, w512patrycja@gmail.com}
}

\maketitle

\begin{abstract}
Emotion detection from faces is one of the machine learning problems needed for human-computer interaction. The variety of methods used is enormous, which motivated an in-depth review of articles and scientific studies. Three of the most interesting and best solutions are selected, followed by the selection of three datasets that stood out for the diversity and number of images in them. The selected neural networks are trained, and then a series of experiments are performed to compare their performance, including testing on different datasets than a model was trained on. This reveals weaknesses in existing solutions, including differences between datasets, unequal levels of difficulty in recognizing certain emotions and the challenges in differentiating between closely related emotions.

\def\thefootnote{*}\footnotetext{These authors contributed equally to this work}
\def\thefootnote{\arabic{footnote}}

\keywords{Facial Emotion Recognition \and Deep learning \and Computer Vision.}
\end{abstract}

\section{Introduction}
This study presents a comprehensive analysis of the most advanced facial emotion recognition solutions currently available. The approach evaluates network performance when trained on one dataset and applied to others.
This methodology enables to ascertain the model's generalisation capabilities, which is a crucial aspect in determining its state-of-the-art status. The study showed that a network's performance on different datasets is significantly worse than when it is tested on the same dataset on which it was trained.

The motivation for this study lies in the crucial role emotions play in decision-making and interpersonal relationships. As technology advances, human-computer systems are expected to evolve, and incorporating emotion recognition is a way for enhancing their responses.
Computer vision methods play a crucial role here as humans perceive messages only in 7\% from the verbal part - words, while as much as 55\% from non-verbal messages - body language and facial expressions \cite{elems_of_emotion}.  By understanding users' emotional states, human - computer interaction systems will be able to engage more naturally, improve user interactions, and foster stronger connections.

A significant gap in current FER research lies in the thorough verification and evaluation of proposed solutions. While many papers claim state-of-the-art performance based on accuracy metrics within their training domain, a comprehensive cross-dataset testing reveals concerning limitations in real-world generalization capabilities. This work addresses this critical need, acknowledging that true progress in FER requires not just innovation, but thorough verification and replication of existing approaches.

\section{Related work}
Universal recognition of facial expressions supports the development of computer vision-based facial emotion recognition (FER) systems, making them effective across different populations. In 1972, Paul Ekman and Wallace V. Friesen \cite{ekman_friesen} conducted a pioneering study identifying six basic emotions: anger, disgust, fear, happiness, sadness, and surprise. By photographing individuals worldwide and asking people from different cultures to match expressions to emotion words in their language, they found strong cross-cultural agreement, suggesting that some facial expressions are universally recognized, even if cultures differ in emotional expressiveness.

Cross-database evaluation is a practice that allows the generalisation capabilities of a model to be tested outside the laboratory environment. It is particularly useful in cases where multiple datasets are available for the same task, such as object recognition\cite{tiny_objects}\cite{object_recognition}  and medical cases\cite{finger_veins}\cite{heartbeat}. Robustness, which is natural for humans, is not so trivial for neural networks, which are vulnerable to noise in the data\cite{humans_vs_non_generalisation} and can also learn not obvious features that interfere with correct predictions\cite{weird_feature_robustness}.

There have been similar reviews\cite{cross-database-2014}\cite{cross-database-2017} of FER methods on different datasets, but they do not include the latest deep learning solutions or the now most popular datasets on which the benchmarks are calculated. Current state-of-the-art models, such as POSTER++\cite{poster_v2}, DMUE\cite{dmue} and DAN \cite{dan} 
were only evaluated on the same test set as they were trained on. These are at the same time models that were chosen for experimentation.

\section{General concept}
Although well-trained models can directly produce high-confidence inferences, the inference results may actually be erroneous. Therefore, evaluating and selecting a model with respect to its robustness to data of different origins is essential. This study examined the basic performance parameters of the model by multi-validation on different datasets.

This work is focused on a discrete representation of emotion, along with a static image-based approach. This choice was driven by the broader availability and greater variety of image-based datasets compared to video-based ones and multiplicity of state-of-the-art models available for discrete problems. We specifically selected "in-the-wild" datasets to better simulate real-world conditions.

These large-scale facial expression recognition datasets often face issues of annotation ambiguity. The challenges of creating reliable labels in static image-based FER datasets are influenced by many factors, such as:
\begin{itemize} 
    \item \textbf{Complexity of emotions} - real-world facial expressions often represent a mixture of emotions, rather than the exaggerated, single emotions typically observed in controlled laboratory settings. Annotators are forced to select a primary emotion (out of 6 basic emotions), potentially misinterpreting the true emotional state.
    
    \item \textbf{Lack of context} - one picture may not provide enough information to distinguish between certain emotions. For example, an expression may be \textit{surprise} or \textit{fear} depending on context. Video-based approaches are a solution to this issue, but they require much more resources.
    
    \item \textbf{Annotators biases} - the ambiguity of facial expressions and the lower quality of images result in different biases in the interpretation of emotions, with a greater inclination towards negativity (anger) rather than positivity (happiness) \cite{annotations}. 
\end{itemize}

These challenges have an impact on the final appearance of the datasets, and the occurrence of internal errors and discrepancies between different datasets.

Creating reliable and robust FER solutions is crucial because errors in emotional recognition can have severe consequences in critical domains like healthcare. Misinterpreting a patient's emotional state could lead to improper mental health diagnoses and potentially harmful treatment decisions. Similarly, false positives identifying emotions that aren't present could trigger unnecessary security responses or interventions.

Furthermore, the cumulative impact of these failures can severely damage organizational credibility and public trust in FER technology, making it essential to prioritize reliability and robustness in development to ensure these systems genuinely benefit society rather than cause unintended harm.

\section{Experiments}

\subsection{FER Models}

The three most effective solutions were selected based on their actual benchmark ranking. These are: POSTER++ \cite{poster_v2}, DMUE\cite{dmue} and DAN\cite{dan}. All of them were declared state of the art in their publications at the time the work was produced. The objective was to train each model using three various datasets, validate them on their respective training datasets, and then multi-validate them on the other datasets.

The initial implementations of the models were obtained from the official repositories provided by the authors. The code was adapted to the hardware capabilities, keeping the original training parameters as close as possible or adjusting them based on the authors' suggestions. Batch size was the parameter we modified most often.

\subsection{Datasets}
As the primary goal of developing a FER system is to use it in real-world application, so datasets collected in uncontrolled environments were selected for analysis. They all employ a single classification method for identifying emotional states:

\begin{enumerate}
\item \textbf{AffectNet}\cite{affectnet}  is the largest existing dataset for FER problem, containing over one million images, while 420,299 out of them are labelled. Due to its size (128GB), a truncated 4GB version created by the authors was used, which contains 291,651 images and has already been resized to 224 x 224 pixels. 
\item \textbf{RAF-DB}\cite{rafdb} -- The Real-world Affective Faces Database is a collection of 29,672 images with average size of 425 × 425 pixels. It is unique from other collections due to its inclusion of age and race.  
\item \textbf{ExpW}\cite{expw} -- Expressions in the Wild is a collection of 91,793 images labelled with 6 basic emotions. Facial area of these images ranges from 23×23 pixels to 180×180 pixels.
\end{enumerate}

All of the datasets include seven classes - six for basic emotions and one \textit{neutral} and contain a representative amount of decent quality images that have the potential to reflect the real world conditions.

\begin{figure}[H]
\centering
\begin{subfigure}{0.3\textwidth}
 \centering
 \includegraphics[width=\linewidth]{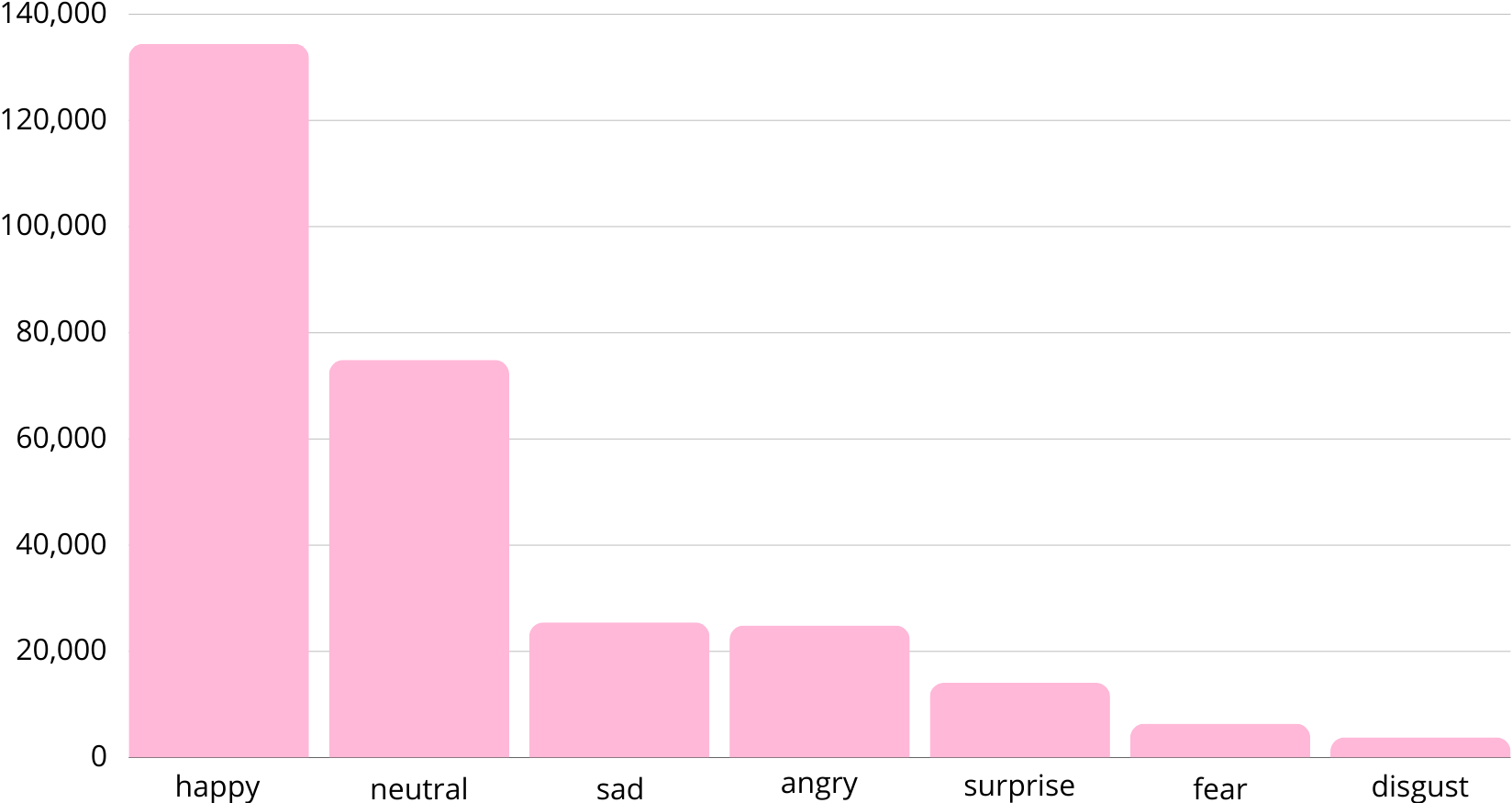}
  \caption{AffectNet}
 \label{fig:sub1}
\end{subfigure}
\begin{subfigure}{0.3\textwidth}
  \centering
  \includegraphics[width=\linewidth]{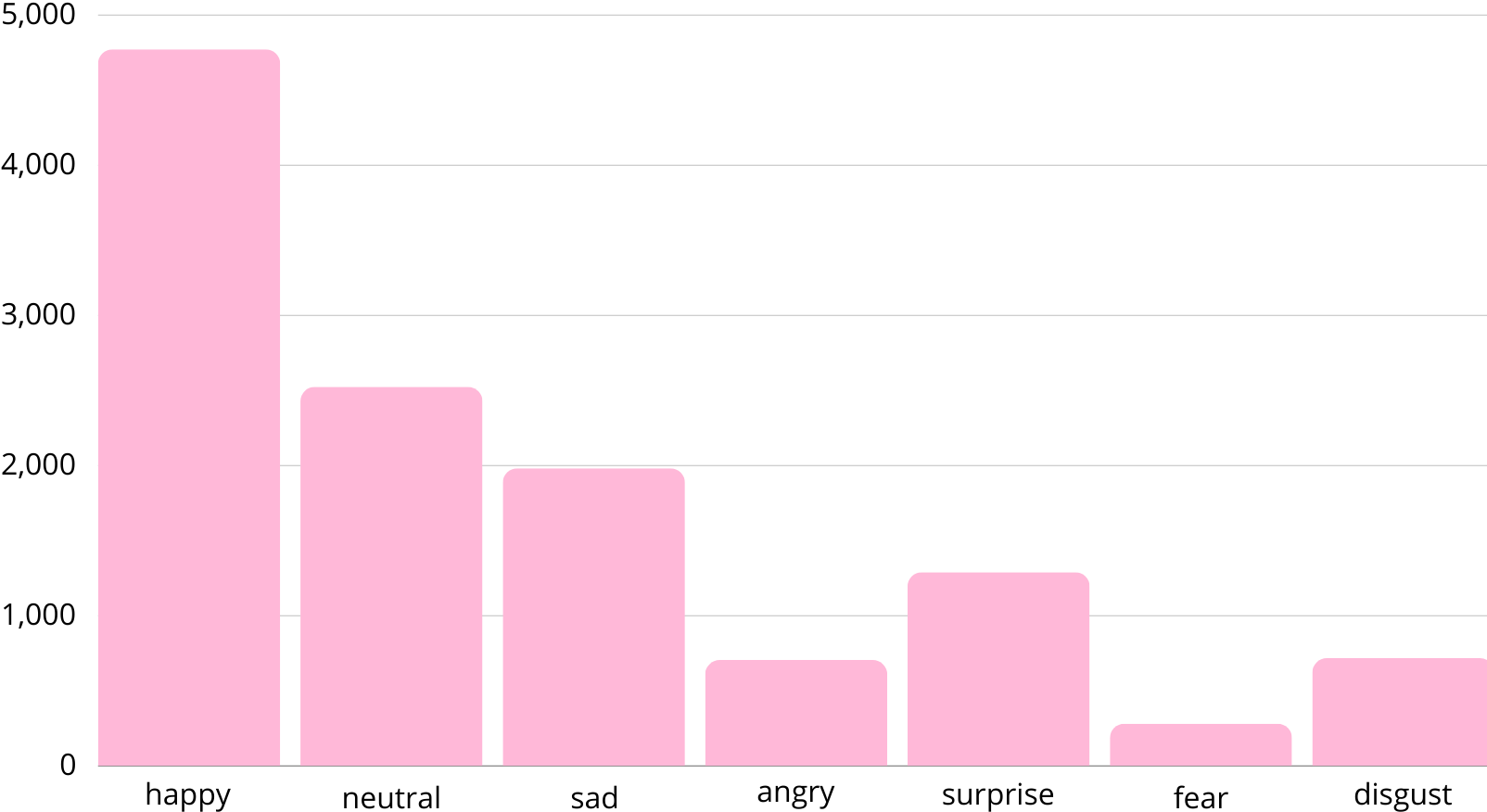}
  \caption{RAF-DB}
  \label{fig:sub2}
\end{subfigure}
\begin{subfigure}{0.3\textwidth}
  \centering
  \includegraphics[width=\linewidth]{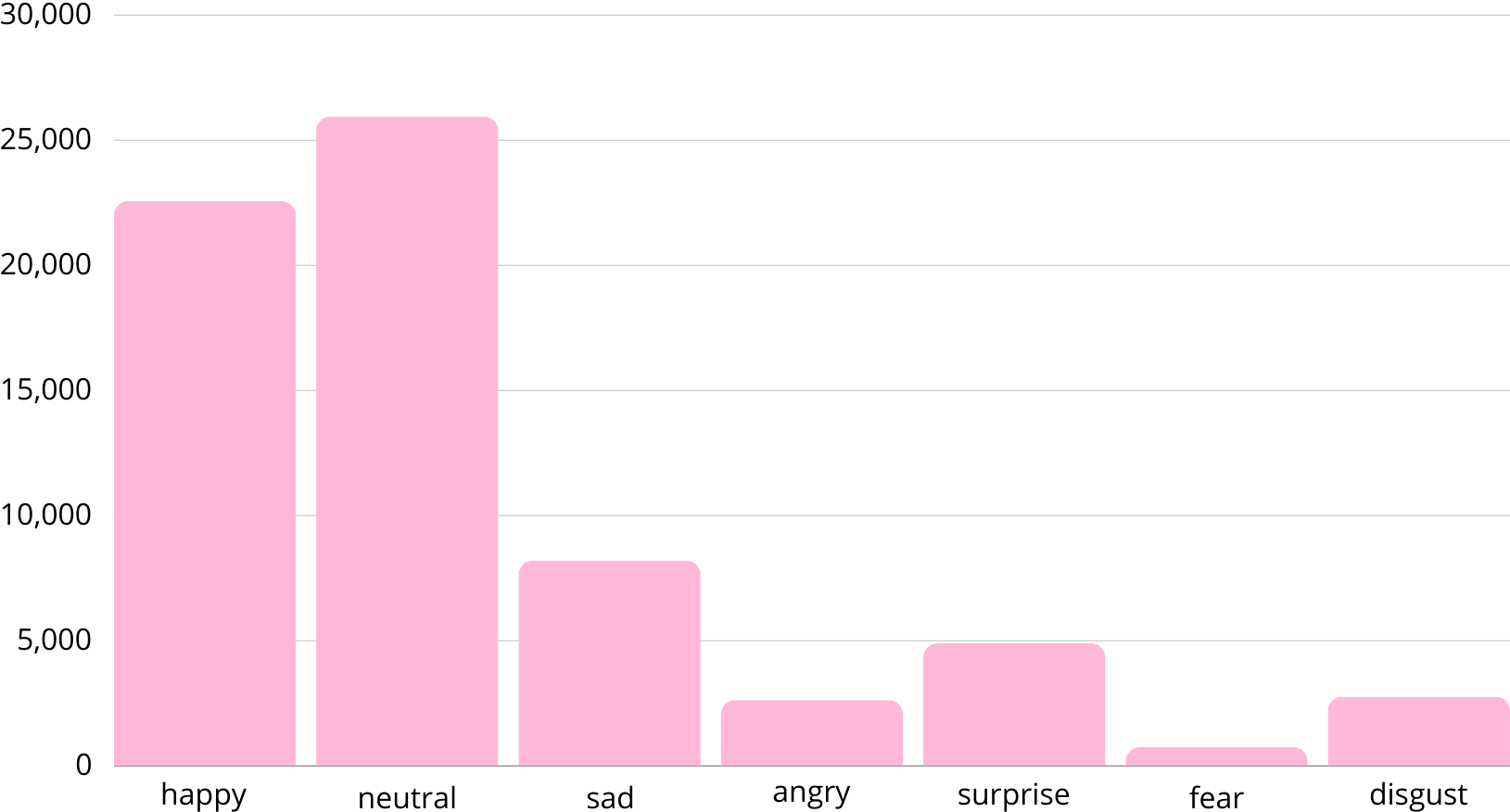}
  \caption{ExpW}
  \label{fig:sub2}
\end{subfigure}

\caption{Class distributions in chosen datasets. Each set contains more images marked as \textit{happy}, with \textit{neutral} coming in second place. All other classes have significantly fewer samples. The collections were created by bulk downloading images from the internet and it is suspected that \textit{happy} and \textit{neutral} images are the most abundant on the web, as they are the images people upload most frequently.}
\label{fig:test}
\end{figure}

\subsection{Training and testing}

The determinant of model quality is most often the accuracy achieved on a test set derived from the same dataset used for training. This enables an evaluation of the extent to which the model has learned images of a given characteristic and facilitates a clear comparison of models with each other. If the model is subsequently deployed in a different context than it was trained, it may result in reduced performance due to the fit to a particular type of image. To minimise this risk and thus identify the actual best model, we opted for multi-validation. The following section presents a summary of the models' performance using accuracy and F1-score.
Accuracy is presented to align the experiments' results with those reported in the original articles. However, it is essential to consider the limitations of this metric. F1-
score is therefore included to provide a more comprehensive assessment of model performance.

\begin{table}[H]

\centering
\begin{threeparttable}
\begin{tabular}{|>{\centering\arraybackslash}p{0.20\textwidth}|
                >{\centering\arraybackslash}p{0.18\textwidth}|
                >{\centering\arraybackslash}p{0.18\textwidth}|
                >{\centering\arraybackslash}p{0.18\textwidth}|
                >{\columncolor[HTML]{EFEFEF}\centering\arraybackslash}p{0.18\textwidth}|}
\hline
\diagbox{Model}{Dataset} & \textbf{AffectNet} & \textbf{RAF-DB} & \textbf{ExpW} & \multicolumn{1}{c|}{\cellcolor[HTML]{EFEFEF}\textbf{Average}} \\ 
\hline
\textbf{POSTER++} & 0.67 / 0.67 & \textbf{0.92 / 0.87} & 0.73 / 0.50 & \textbf{0.77 / 0.68} \\ 
\hline
\textbf{DAN} & 0.62 / 0.62 & 0.76 / 0.68 & 0.60 / 0.44 & 0.60 / 0.58 \\ 
\hline
\textbf{DMUE} & 0.62 / 0.62 & 0.80 / 0.73 & 0.65 / 0.49 & 0.69 / 0.58 \\ 
\hline
\rowcolor[HTML]{EFEFEF} \textbf{Average} & 0.64 / 0.64 & \textbf{0.83 / 0.76} & 0.66 / 0.48 &  \\ 
\hline
\end{tabular}

\end{threeparttable}
\vspace{10pt}
\caption{Accuracy / F1-Score Comparison Across Datasets. Some of these results are slightly lower from the ones obtained by their authors, but the differences are overall negligible in the context of further experiments.}
\label{tab:accuracy_f1}
\end{table}

The results of the experiments demonstrate that each model exhibited a decline in performance when evaluated on a dataset that differed from the one used for training. The tables 2-4 present a comprehensive overview of the outcomes for all models and tests.


\begin{table}[H]
\centering
\begin{tabular}{|>{\centering\arraybackslash}p{0.26\textwidth}|
                >{\centering\arraybackslash}p{0.15\textwidth}|
                >{\centering\arraybackslash}p{0.15\textwidth}|
                >{\centering\arraybackslash}p{0.15\textwidth}|
                >{\columncolor[HTML]{EFEFEF}\centering\arraybackslash}p{0.15\textwidth}|}
\hline
\diagbox{Trained on}{Tested on} & \textbf{AffectNet} & \textbf{RAF-DB} & \textbf{ExpW} & \textbf{Average} \\ 
\hline
\textbf{AffectNet}              & 0.67 / 0.67        & 0.75 / 0.63     & 0.54 / 0.39   & \cellcolor[HTML]{EFEFEF}\textbf{0.65} / 0.56 \\ 
\hline
\textbf{RAF-DB}                 & 0.49 / 0.47        & \textbf{0.92 / 0.87} & 0.52 / 0.39   & \cellcolor[HTML]{EFEFEF}0.64 / \textbf{0.58} \\ 
\hline
\textbf{ExpW}                   & 0.42 / 0.36        & 0.77 / 0.59     & 0.73 / 0.50   & \cellcolor[HTML]{EFEFEF}0.64 / 0.48 \\ 
\hline
\cellcolor[HTML]{EFEFEF}\textbf{Average} & \cellcolor[HTML]{EFEFEF}0.53 / 0.50 & \cellcolor[HTML]{EFEFEF}\textbf{0.81 / 0.70} & \cellcolor[HTML]{EFEFEF}0.60 / 0.43 & \cellcolor[HTML]{EFEFEF} \\ 
\hline
\end{tabular}
\vspace{10pt}
\caption{POSTER++ \textbf{accuracy} / \textbf{macro average F1 score}}
\label{tab:poster_combined}
\end{table}

\vspace{-10pt}


\begin{table}[H]
\centering
\begin{tabular}{|>{\centering\arraybackslash}p{0.26\textwidth}|
                >{\centering\arraybackslash}p{0.15\textwidth}|
                >{\centering\arraybackslash}p{0.15\textwidth}|
                >{\centering\arraybackslash}p{0.15\textwidth}|
                >{\columncolor[HTML]{EFEFEF}\centering\arraybackslash}p{0.15\textwidth}|}
\hline
\diagbox{Trained on}{Tested on} 
                                                  & \textbf{AffectNet}           & \textbf{RAF-DB}                       & \textbf{ExpW}                & \textbf{Average} \\ 
\hline
\textbf{AffectNet}                                & 0.62 / 0.62                 & 0.65 / 0.53                            & 0.46 / 0.34                   & \cellcolor[HTML]{EFEFEF}\textbf{0.58 / 0.50} \\ 
\hline
\textbf{RAF-DB}                                   & 0.40 / 0.39                 & \textbf{0.76 / 0.68}                  & 0.37 / 0.28                   & \cellcolor[HTML]{EFEFEF}0.51 / 0.45 \\ 
\hline
\textbf{ExpW}                                     & 0.44 / 0.43                 & 0.65 / 0.52                            & 0.60 / 0.44                   & \cellcolor[HTML]{EFEFEF}0.56 / 0.46 \\ 
\hline
\cellcolor[HTML]{EFEFEF}\textbf{Average}         & \cellcolor[HTML]{EFEFEF}0.49 / 0.48 & \cellcolor[HTML]{EFEFEF}\textbf{0.69 / 0.58} & \cellcolor[HTML]{EFEFEF}0.48 / 0.35 & \cellcolor[HTML]{EFEFEF} \\ 
\hline
\end{tabular}
\vspace{10pt}
\caption{DAN \textbf{accuracy} / \textbf{macro average F1 score}}
\label{tab:dan_combined}
\end{table}



\begin{table}[H]
\centering
\begin{tabular}{|>{\centering\arraybackslash}p{0.26\textwidth}|
                >{\centering\arraybackslash}p{0.15\textwidth}|
                >{\centering\arraybackslash}p{0.15\textwidth}|
                >{\centering\arraybackslash}p{0.15\textwidth}|
                >{\columncolor[HTML]{EFEFEF}\centering\arraybackslash}p{0.15\textwidth}|}
\hline
\diagbox{Trained on}{Tested on} 
                                                  & \textbf{AffectNet}           & \textbf{RAF-DB}              & \textbf{ExpW}                & \textbf{Average} \\ 
\hline
\textbf{AffectNet}                                & 0.62 / 0.62                 & 0.62 / 0.53                 & 0.39 / 0.31                  & \cellcolor[HTML]{EFEFEF}0.54 / 0.49 \\ 
\hline
\textbf{RAF-DB}                                   & 0.45 / 0.44                 & \textbf{0.80 / 0.73}        & 0.47 / 0.35                  & \cellcolor[HTML]{EFEFEF}0.57 / 0.51 \\ 
\hline
\textbf{ExpW}                                     & 0.48 / 0.46                 & 0.75 / 0.60                 & 0.65 / 0.49                  & \cellcolor[HTML]{EFEFEF}\textbf{0.63 / 0.52} \\ 
\hline
\cellcolor[HTML]{EFEFEF}\textbf{Average}         & \cellcolor[HTML]{EFEFEF}0.52 / 0.51 & \cellcolor[HTML]{EFEFEF}\textbf{0.74 / 0.62} & \cellcolor[HTML]{EFEFEF}0.50 / 0.38 & \cellcolor[HTML]{EFEFEF} \\ 
\hline
\end{tabular}
\vspace{10pt}
\caption{DMUE \textbf{accuracy} / \textbf{macro average F1 score}}
\label{tab:dmue_combined}
\end{table}

Testing on the same dataset used for training yields the best results; any changes to this dataset lead to a decline in performance metrics. Some decreases are minor; for instance, the Poster model trained on AffectNet shows a variation of about ±10 percentage points when tested on other datasets. However, certain drops are drastic, such as the DAN model trained on RAF-DB, which achieves 76\% accuracy on RAF-DB but only 37\% on ExpW. Additionally, the accuracy metric significantly outperforms the F1 score, indicating an issue with class imbalance and poor performance on certain classes. Using accuracy as a main metric by the authors of state of the art solutions covers up their imperfections.


\begin{figure}[H]
\centering
\begin{subfigure}{0.5\textwidth}
 \centering
  \includegraphics[width=\linewidth]{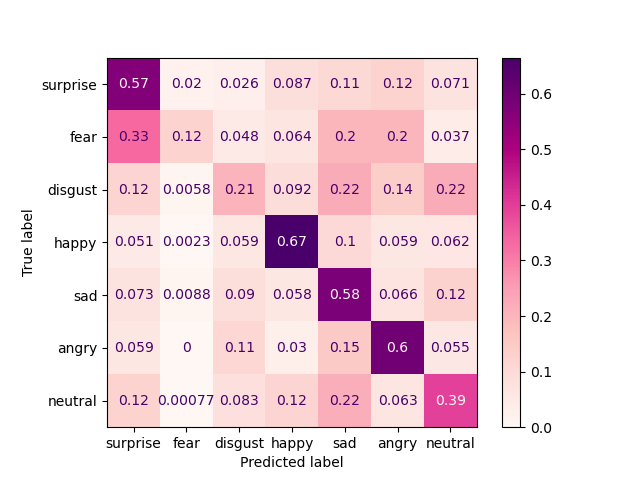}
  \caption{Training dataset: RAF-DB}
 \label{fig:sub1}
\end{subfigure}%
\begin{subfigure}{0.5\textwidth}
  \centering
   \includegraphics[width=\linewidth]{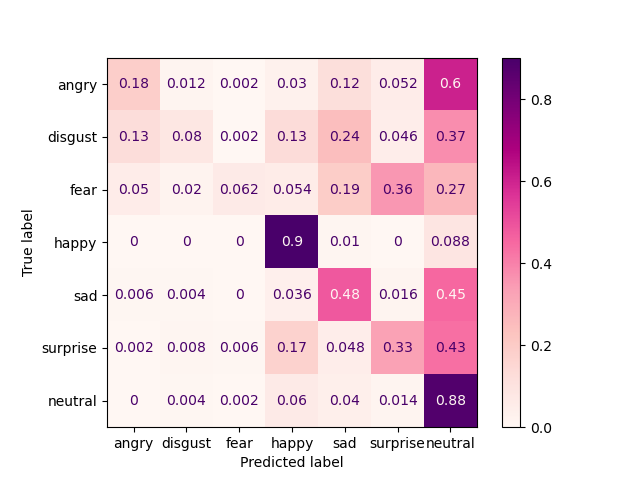}
  \caption{Training dataset: ExpW}
  \label{fig:sub2}
\end{subfigure}
\caption{Results of POSTER++ models evaluation on AffectNet dataset. These highlight differences between the datasets, exposing classes that are confused with each other most often. A) The most confused pairing between RAF-DB and AffectNet is surprise with fear, and sad with neutral and disgust. B) There are a great deal of images labelled neutral in ExpW, which has resulted in over-classification as neutral, rather than more varied results. It can also be seen that the least frequent classes in ExpW (angry, disgust and fear) have drastically the worst results. }
\label{fig:test}
\end{figure}

\section{Conclusions}

The conclusion drawn from this is that the average face expressing a particular emotion varies across datasets. Each dataset was labelled independently, introducing biases based on the labelling conventions adopted by its creators. Emotions can be subjectively interpreted by different individuals and are given subjective labels. For example \textit{happy} is generally recognized accurately by most people, being the easiest emotion to identify without confusion. Creating a good and objectively labelled dataset is a challenging task, given the fact that 'in the wild' datasets are annotated only after data collection. This makes it impossible to be certain whether a person actually experienced a particular emotion, but it does provide a basis for developing a model that can work in real-world settings where faces differ and do not resemble those in the dataset.

\subsubsection{Minority classes}
The models have difficulties in recognising minority classes such as fear, sadness and disgust, performing with greater accuracy for happy expressions, despite efforts to achieve a balanced distribution of classes. This is a result of an overfitting to the more prevalent happy class, combined with inherent similarities between some emotional expressions (surprise and fear) that can even mislead human annotators.

Improvements to address this issue could focus on increasing the representation of minority classes, involving more annotators to reduce bias, and filtering out images with low confidence labels to minimize the impact of uncertain annotations on model performance.

\subsubsection{Datasets differences}
Models show notable results when tested on samples from the same dataset as the training data. However, they struggle with generalisation on multi-validation datasets, which clearly demonstrates the challenges in transferring knowledge across datasets. Notably, AffectNet and ExpW models show good results on RAF-DB, which indicates that they share some characteristics with RAF-DB but that there is limited overlap between AffectNet and ExpW.

\subsubsection{Truth about accuracy}
Many academic works emphasize accuracy as the primary indicator of a model’s success, often claiming state-of-the-art status based solely on this metric. However, focusing on accuracy alone can overlook critical aspects like precision in minority classes, potentially masking a model's true effectiveness, especially in nuanced emotion recognition. 

This issue becomes particularly evident when examining F1 scores, which frequently reveal significantly lower performance than accuracy metrics, indicating that models may excel at identifying majority classes while struggling with less represented emotions, thus painting a misleading picture of their real-world applicability.

\subsubsection{Possible improvements}

Enhancing FER systems requires a multi-faceted approach starting with the foundation of high-quality datasets that are carefully balanced across different races, genders, ages, and cultural expressions to reduce bias and improve generalization. The development of such datasets would particularly benefit from close collaboration between AI researchers and psychologists, as their expertise in human emotion can ensure more accurate labeling and better representation of subtle emotional states. Another training techniques, like semi-supervised learning approaches (eg. Noisy Student Training \cite{noisystudent}), can significantly improve robustness and accuracy in real-world conditions. For maximum reliability, FER could be integrated into multimodal emotion recognition systems that combine facial analysis with voice emotion detection, as this fusion of different data sources can provide more accurate and context-aware emotional assessments than any single modality alone.

%
%
\bibliographystyle{splncs04}
%






\end{document}